\documentclass[runningheads]{llncs}
\usepackage[T1]{fontenc}
\usepackage{graphicx}
\usepackage{pifont}    

\usepackage{booktabs}

\usepackage[hidelinks]{hyperref}

\usepackage{amsmath}

\newcommand{\argmin}{\arg\!\min}
\newcommand{\argmax}{\arg\!\max}

\usepackage{dsfont}

\usepackage{subcaption}

\usepackage{makecell}

\usepackage{overpic}

\usepackage{amssymb}

\begin{document}
\title{Choosing a parallel heterogeneous ensemble method for tabular classification}
%
\titlerunning{Choosing a parallel heterogeneous ensemble method}
%
\author{Vassili Maillet\inst{1, 2}\\\orcidID{0009-0001-7827-0989}  \\
Gustavo (Jesús) Angulo\inst{1}\\ \orcidID{0000-0001-6106-2692} \\
Pierre Jouvelot\inst{1}\\ \orcidID{0000-0002-6783-5796}}
\authorrunning{V. Maillet et al.}
%
\institute{Mines Paris, PSL University, France, \\ \email{vassili.maillet@minesparis.psl.eu} \\
\and Alten Labs, France}
\maketitle              
\begin{abstract}

\par  Parallel ensemble methods were compared on $56$ small-to-medium tabular classification tasks drawn from OpenML CC18. A set of ``best practice'' recommendations on the use of ensemble methods was derived from these observations. It was later validated on 28 additional tasks using TabArena's precomputed data, where the recommendation set significantly outperformed Single Best and matched or exceeded individual ensemble methods.

\par Two key observations were made. First, Blending and Stacking are inconsistent, but their inconsistencies are independent and happen on different tasks. Second, while Hard Voting's probabilistic classification is rather weak, a consequence of using vote proportions as posterior estimates, Robust Soft Voting's probabilistic classification is particularly successful, especially in the multiclass case.






\end{abstract}
%
%
%

\section{Introduction}

\par Ensemble methods improve the performance of individual models, called \textit{base learners}, by combining them. The set of base learners of an ensemble model, its \textit{pool}, is \textit{homogeneous} if every base learner is of the same family, and \textit{heterogeneous} otherwise. Homogeneous tree-based ensemble models have been considered some of the best performing models in tabular classification for years~\cite{grinsztajn_why_2022}. If much has been written about their performance, the capabilities of heterogeneous ensemble models' have been less studied (see Section~\ref{sec:related}), despite their prevalence in machine-learning competitions such as on Kaggle.
\vspace{8pt}
\par In this work, we provide the following contributions: 
\vspace{-8pt}
\begin{itemize}
    \item a thorough experimental study of the performances of 9 different parallel ensemble methods using a diverse and heterogeneous pool;
    \item a new set of ``best practice'' recommendations on the use of such ensemble methods, for small to large datasets;
    \item a validation of these recommendations using TabArena's precomputed data~\cite{erickson_tabarena_2025};
    \item and two key observations, namely (1) that Blending and Stacking are inconsistent, but their inconsistencies are independent and happen on different tasks and (2), while Hard Voting's probabilistic classification is rather weak, Robust Soft Voting's own probabilistic classification is particularly successful, especially in the multiclass case.
\end{itemize}

\par The structure of the paper is as follows. After surveying the related work (Section~\ref{sec:related}), we evaluate ensemble methods in a first exploratory study on a simple pool (Section~\ref{sec:exploratory}). Then, we make recommendations (Section~\ref{sec:analysis}) that we validate in a second study (Section~\ref{sec:validation}), closer to real-life conditions, using TabArena's pre-computed results, before concluding (Section~\ref{sec:conclusion}).

\section{Related Work}
\label{sec:related}

\par Previous studies have often focused on homogeneous ensemble models~\cite{opitz_popular_1999}~\cite{cruz_dynamic_2018}, but works comparing heterogeneous ensemble models do exist. Creators of ensemble methods and AutoML researchers interested in post-hoc ensembling have compared the results of different ensemble methods in different ways.

\par In the first case, studies have focused on a specific subset of ensemble methods, such as Stacking~\cite{dzeroski_is_2004}, or were done using a few datasets~\cite{caruana_getting_2006}. The work most similar to ours' is~\cite{dzeroski_is_2004}, which compared many types of Stacking techniques with both Hard Voting and Single Best on $30$ different datasets. Our work differs in terms of ensemble methods, pools, datasets, and objective, as we have access to more modern classifiers, test a larger curated set of datasets~\cite{bischl_openml_2021}\cite{erickson_tabarena_2025} and don't focus only on Stacking. 

\par In the second case, the comparison of different ensemble methods was often not the focus. The closest work to ours lies in the appendix of the Assembled OpenML text~\cite{purucker_assembled-openml_2023}, which provided a quick comparison of different ensemble methods based on base learners found by a CASH method, optimizing both the choice of the family of learner as well as their hyperparameters. This represented well the results of an AutoML method but left in some cases a homogeneous pool. In~\cite{feurer_efficient_2015}, Ensemble Selection~\cite{caruana_getting_2006} and Stacking were quickly compared in the context of post-hoc ensembling with the first being considered faster and more robust. Recently,~\cite{xu_pseo_2025} offered a comparison of several modern post-hoc ensembling methods. Our approach deals with a wider spectrum of ensemble methods, while the resulting recommendations have been shown to generalize to large datasets.

\par Taking together both perspectives, prior comparisons either focused on a restricted set of methods, typically Stacking versus Hard Voting, or used homogeneous pools, outdated base learners, or small dataset collections. Our work extends this landscape in three ways: (1) we evaluate a broader set of parallel ensemble methods, including cluster-based dynamic selection and the median aggregation rule; (2) we use a deliberately diverse heterogeneous pool that includes modern classifiers (XGBoost, multiple MLP architectures); and (3), more relevant for the end-user perspective, we translate empirical observations into a concrete decision procedure and validate it on an independent benchmark. To our knowledge, no prior work simultaneously provides recommendations and validates them empirically on a separate benchmark suite.



\section{Exploratory Experiments}
\label{sec:exploratory}

We select a set of widely used heterogeneous ensemble methods and apply them on tasks from OpenML CC18~\cite{bischl_openml_2021}. 

\subsection{Ensemble Methods and Base Learners}


\subsubsection{Ensemble Methods}

\par Our approach assumes that no estimation of the performance of base learners on the given tasks is available. This differs from cases such as post-hoc ensembling~\cite{feurer_efficient_2015} where base learners are first trained, evaluated, and then selected for an ensemble based on their evaluation. We also focus on ensemble methods whose base learners' training is independent, to allow a base learner to be reused across ensemble methods with the same training scheme.

\par We studied three groups of model-agnostic ensemble methods: voting, stacked generalization and dynamic selection.
\begin{itemize}
    \item We chose Soft Voting~\cite{dietterich_ensemble_2000}, Hard Voting~\cite{dietterich_ensemble_2000} and Bagging~\cite{breiman_bagging_1996} because of their wide use. For Hard Voting's probabilistic classification, we use the proportion of voters for each class, a known yet rarely used technique. We also added Robust Soft Voting - the median rule - for its robustness.



    \item We also added Stacking~\cite{wolpert_stacked_1992} and Blending~\cite{brownlee_blending_2020} for their popularity. Both use base learners' predictions to generate a meta-dataset on which they train a final estimator, a logistic regression in our case. Their difference lies in the base learners' training. In Blending, they are trained once through a $90/10$ training/holdout scheme. In Stacking, base learners follow a $5$-fold cross validation to create the meta-dataset from out of fold predictions, then the base learners used for predictions are trained on the whole training set.

    \item Finally, we included a Dynamic Classifier Selection (DCS) method and a Mixture of Experts (MoE) method, which part the feature space using clustering techniques.
\end{itemize}

\par Details about the methods and their implementation may be found in Appendix A. 

\subsubsection{Base Learners}

\par We chose a diverse set of well-known classifiers with different inductive biases, representative of typical practical uses: two Decision Trees (one with a limited depth, the other, without), two SVM (with  linear and RBF kernels), a very simple model giving the Most Common class, a Gaussian Naïve Bayes model, a Logistic Regression, a Random Forest, a XGBoost, and three MLPs (one with the default size ($5 \times 10$), one with only one long hidden layer ($1 \times 100$) and one with four times the size of the default ($10 \times 20$)). 

\par These models come from the \texttt{scikit-learn} library~\cite{pedregosa_scikit-learn_2011}, with the exception of the popular XGBoost~\cite{chen_xgboost_2016} and Most Common. In most cases, we kept the default values instead of biasing the results with the authors' choice of hyperparameter values. We only chose different hyperparameter values to avoid overly long training times, and to make variations, if we expected them to have a major impact on the results. 

\par Each dataset was preprocessed as follows: (1) missing features were imputed; (2) we used a one-hot encoder on categorical features and dropped constant features; and (3) each feature's mean and variance were scaled between $0$ and $1$.

\par Implementation details may be found in Appendix B.

\subsection{Methodology}

\par The performances of ensemble methods were measured using mainly their mean ROC AUC OVO from five repetitions of a $10$-fold cross validation.

\par As mentioned above, we used the OpenML CC18~\cite{bischl_openml_2021} suite of classification tasks, but removed the $16$ biggest datasets - those with more than $500,000$ cells - due to limited computational resources. This left us with $28$ binary tasks and $28$ multiclass tasks.

\subsection{Results}
\label{sec:results}

\subsubsection{General Results}

\par \autoref{tab:basic_metrics} shows the mean score (from low, 0, to high, 1), mean rank (lower is better) and ratio of victories (being the best method) of the ensemble methods for the ROC AUC and two classification metrics, Matthew's Correlation Coefficient (MCC) and  accuracy.\footnote{Measures and further details may be found at \doi{10.5281/zenodo.17986269}.}

\begin{table}[ht]
	\centering
	\caption{Mean scores, mean ranks, task-wise rank Mean Absolute Deviation (MeAD) and win ratio of the ensemble methods over all tasks for different scores. Best results are underlined.}
	\label{tab:basic_metrics}
    \begin{tabular}{rccccccccc}
    	\hline                 
    	&  \multicolumn{3}{c}{\textbf{ROC AUC}}  &  \multicolumn{3}{c}{\textbf{MCC}}   &  \multicolumn{3}{c}{\textbf{Accuracy}}  \\
    	&   \hspace{1mm}Rank\hspace{1mm}  &  \hspace{1mm}Score\hspace{1mm}   & Wins &   \hspace{1mm}Rank\hspace{1mm}  &  \hspace{1mm}Score\hspace{1mm}  & Wins  &   \hspace{1mm}Rank\hspace{1mm}  &   \hspace{1mm}Score\hspace{1mm}  & Wins \\
    	\hline
        \textbf{Single Best} &  $4.4 \pm 2.0$   &     0.922  & $16\%$    & $4.4 \pm 2.0$ &   0.714 & $13\%$    &   $4.8 \pm 2.2$   &      \underline{0.872}  &  $9\%$   \\
        \hline
        \textbf{Robust SV}  &  $\underline{2.8} \pm 1.5$   &     \underline{0.926} & $31\%$      & $4.5 \pm 1.9$ &   0.705  & $9\%$   &   $4.0 \pm 2.0$   &      0.869   &  $17\%$  \\
        \textbf{Stacking}   &  $3.0 \pm 1.9$   &     0.921  & \underline{$38\%$}     & $\underline{3.3} \pm 2.8$ &   0.710  & \underline{$53\%$}   &   $\underline{3.1} \pm 2.4$   &      0.871   &  \underline{$48\%$} \\
        \textbf{Soft Voting} &  $3.7 \pm 1.1$   &     0.925   & $2\%$    & $4.4 \pm 1.5$ &   0.708  & $4\%$   &   $4.3 \pm 1.4$   &      0.870   &  $4\%$ \\
        \textbf{Blending}   &  $3.8 \pm 1.5$   &     0.924  & $9\%$     & $3.9 \pm 1.9$ &   \underline{0.720}  & $11\%$   &   $3.6 \pm 1.7$   &      0.879   &  $12\%$ \\
        \textbf{Bagging}   &  $5.3 \pm 1.0$   &     0.921  & $0\%$     & $6.4 \pm 1.3$ &   0.699 & $2\%$    &   $6.4 \pm 1.1$   &      0.865   &  $0\%$ \\
        \textbf{Cluster DCS} &  $6.7 \pm 1.1$   &     0.911  & $2\%$     & $5.6 \pm 1.9$ &   0.713  & $1\%$   &   $5.9 \pm 1.8$   &      0.870    & $4\%$  \\
        \textbf{Hard Voting} &  $7.2 \pm 1.4$   &     0.907  & $0\%$     & $5.4 \pm 2.2$ &   0.697  & $3\%$   &   $5.0 \pm 2.0$   &      0.869    & $4\%$ \\
        \textbf{Cluster MoE} &  $8.2 \pm 1.0$   &     0.900   & $2\%$    & $7.1 \pm 1.9$ &   0.693  & $4\%$   &   $7.8 \pm 1.4$   &      0.860   &  $2\%$ \\
        \hline
    \end{tabular}
\end{table}




\par \autoref{tab:basic_metrics} reveals a clear metric-dependent split. In probabilistic classification (ROC AUC), 
Robust SV leads, followed by Soft Voting and Stacking; the spread across methods is modest ($0.026$ between best and worst). 
In discriminative classification (MCC, accuracy), Stacking and Blending lead, while Robust SV drops to rank $4.5$ and $4$. 
This divergence arises because the median aggregation rule smooths probability estimates toward the center of the simplex, which benefits calibration and AUC but can hurt hard-decision boundaries. 
Hard Voting performs poorly in probabilistic classification because its class probabilities are estimated by the vote proportion, a coarse approximation of the true posterior\cite{hastie_elements_2009}.

\subsubsection{Specific-Task Results}

\par
\begin{figure}[ht]
	\begin{subfigure}[t]{\textwidth}
		\centering
		\includegraphics[width=\linewidth]{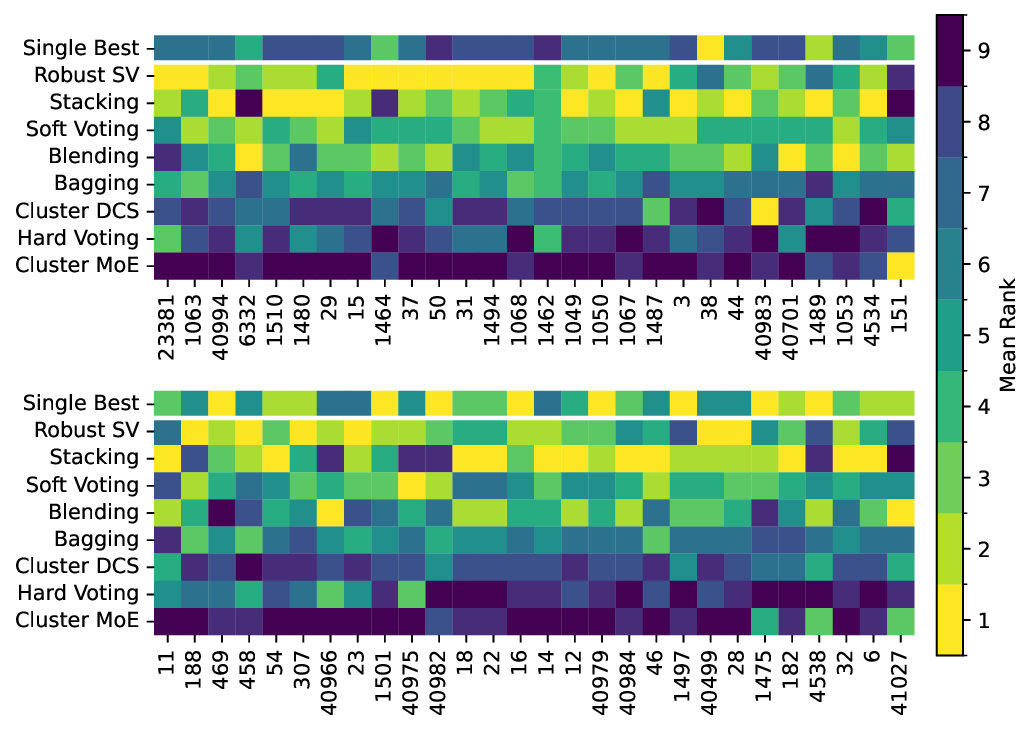}
	\end{subfigure}
	\caption{Heat maps of ROC AUC mean ranks for all datasets (binary tasks on top, multiclass on bottom).}
	\label{fig:heatmaps}
\end{figure}

 We make a more fine-grained study of the models (see \autoref{fig:heatmaps}), using heat maps of the ROC AUC OVO mean rank of each ensemble method for each dataset. Datasets are referenced by their OpenML ID~\cite{bischl_openml_2025}, and ordered, in the figure, according to their number of samples (smaller to larger from left to right).

\par First, we see that Stacking and Blending have independent consistency problems. While having the best overall performances, each of them is one of the two worst methods in $11$ out of $56$ tasks. Robust SV and Soft Voting are more consistent. 

\begin{table}[ht]
\centering
\caption{ROC AUC OVO mean rank among datasets with less than $2,000$ samples, more than $2,000$ samples, binary tasks and multiclass tasks.}
\label{tab:difference_parts}
\begin{tabular}{rcccc}

	\hline
	& \textbf{Small} & \textbf{Medium} &  \textbf{Binary} & \textbf{Multiclass}  \\
	\hline
	\textbf{Single Best} & 5.1 & 3.7 & 5.8 & \underline{3.0} \\
	\hline
    \textbf{Robust SV} & \underline{1.9} & 3.8 & \underline{2.5} & 3.1 \\
    \textbf{Stacking} & 3.5 & \underline{2.4} & 2.8 & 3.1 \\
    \textbf{Soft Voting} & 3.4 & 3.9 & 3.3 & 4.0 \\
    \textbf{Blending} & 4.3 & 3.3 & 3.5 & 4.1 \\
    \textbf{Bagging} & 4.8 & 5.8 & 5.1 & 5.5 \\
    \textbf{Cluster DCS} & 7.0 & 6.4 & 6.7 & 6.8 \\
    \textbf{Hard Voting} & 6.3 & 8.1 & 7.1 & 7.2 \\
    \textbf{Cluster MoE} & 8.7 & 7.6 & 8.3 & 8.0 \\
    \hline
    \end{tabular}
\end{table}

\par Secondly, results change as datasets increase in size. We see this better in \autoref{tab:difference_parts}. Voting-based methods relatively worsen as datasets become larger, while other methods improve. 


\subsubsection{Importance of Base Learners}





\par
\begin{figure}[ht]
    \centering
    \includegraphics[width=\linewidth]{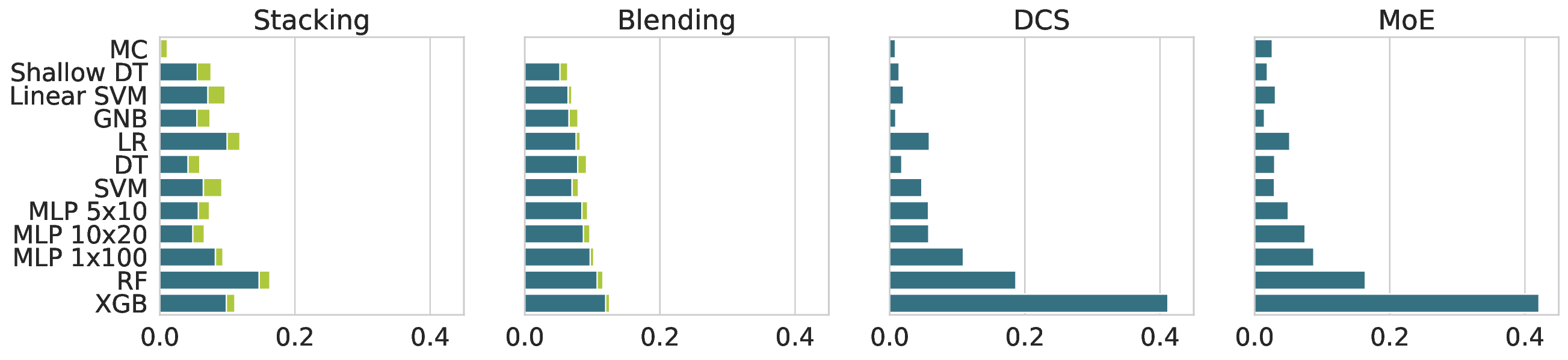}
    \caption{Importance of each base learner in the ensemble method.}
	\label{fig:component_use}
\end{figure}

 While voting methods give the same importance to every base learner, others do not. Stacked generalization methods' final estimators give a different weight to each base learner while selection methods only select one base learner for each sample. We show in \autoref{fig:component_use} a measure of the importance given to each base learner by each ensemble method, a ratio between $0$ and $1
$. For stacked generalization, we use the average absolute weight given to each base learner by the final estimator. For selection methods, we use the percent of samples for which a base learner was selected over the others.

\par In the first case, correlated base learners' results could lead to arbitrarily high weights. We thus monitor the proportion of weights with an illogical sign, such as a negative weight for the probability of class A given by a base learner when computing the probability of class A by the ensemble model. They are in light green in \autoref{fig:component_use}.





\section{Analysis}
\label{sec:analysis}

\subsection{Stacking}

\par Stacking has the best classification performance (\autoref{tab:basic_metrics}) and is able to ignore lower-performing base learners (\autoref{fig:component_use}). However, its training scheme is time-consuming and different from all the ensemble methods studied here, meaning most of its base learners aren't usable by other ensemble methods. Finally, while its results are usually good, they are inconsistent, the method performing badly on some tasks (\autoref{fig:heatmaps}).

\par Blending is fairly similar. It has the second best classification performance (\autoref{tab:basic_metrics}) and the same ability to ignore the worst base learners (\autoref{fig:component_use}). It performs slightly worse on smaller datasets, likely not having enough samples to train its final estimator. It's faster to train, and its results are slightly less inconsistent (\autoref{fig:heatmaps}). Its base learners' simple training scheme allows other ensemble methods to opportunistically use them.

\par Interestingly, while Stacking and Blending both occasionally performed badly, with every repetition showing the same low performance, they didn't do so on the same tasks. Their inconsistencies appeared independent, which implies they don't come from the base learners, but from the final estimator's training.

\subsection{Voting}

\par Voting methods performed well on smaller datasets, but their performance decreased as the size of datasets increased (\autoref{tab:difference_parts}). 


\par This behavior is consistent with a bias-variance analysis of ensemble methods~\cite{domingos_unified_2000}.
In voting ensembles, all base learners contribute equally to the final prediction regardless of their individual quality. On small datasets, variance dominates for most learners and uniform averaging provides genuine variance reduction. As dataset size grows, the bias of weaker learners (e.g., Shallow Decision Tree, Most Common) becomes the limiting factor, their error is irreducible given the model class and continues to be weighted equally with stronger learners. Methods with learned weights (Stacking, Blending) and those performing implicit selection (Cluster DCS, MoE) are less susceptible to this regime change, though they introduce their own instabilities. Pruning the pool to retain only competitive base learners would be expected to restore the advantage of voting methods on larger datasets, a direction which could be explored in future work.


\subsection{Dynamic Selection Methods}
\par Cluster DCS and MoE didn't perform well (\autoref{tab:basic_metrics}), each only being the best method in one case (\autoref{fig:heatmaps}). Though they are able to ignore the worse base learners, they have a tendency to overly rely on a single model in many parts of the feature space (\autoref{fig:component_use}). Of note, the single time Cluster MoE was the best method, it used XGBoost for $99.6\%$ of its predictions.

\subsection{Recommendations}

\par Based on the previous  experiment and analysis, we provide evidence-based, easy-to-use recommendations regarding which ensemble method and training scheme to use in practice (see Table~\ref{tab:reco}). They can all be read and used simultaneously, enabling people using multiple methods to either choose the best-performing one after testing them or to combine their results.

\begin{table}
\centering
\caption{Recommendations for ensemble methods.}
\label{tab:reco}
\begin{tabular}{rl}
\hline
1. & If training time is not a concern, use Stacking.\\
2. & If Stacking or Blending is used, use at least one other method.\\
3. &  If base learners may only be trained once, use a $90|10$ holdout scheme, allowing \\ 
  & the use of
   Blending, Robust SV, Soft Voting and Hard Voting.\\
4. & If a dataset has less than $5,000$ samples, use Robust SV.\\
5. & If a dataset is multiclass, use Single Best (but see also Section~\ref{sec:validation}).\\
6. & If a dataset has more than $500$ samples and is binary, use Blending.\\
7. & If a dataset has more than $2,000$ samples and is multiclass, use Blending. \\
\hline
\end{tabular}
\end{table}


\section{Validation on TabArena}
\label{sec:validation}

\par We validate our recommendations through a set of experiments using TabArena's precomputed results~\cite{erickson_tabarena_2025}.






\subsection{Methodology}

\par TabArena provides the predicted probabilities of many modern classifiers for each sample of a curated list of datasets~\cite{erickson_tabarena_2025}. It's thus possible to simulate ensembling these classifiers as well as running AutoML optimizers. Predictions come from the repetitions of $3$-fold cross validations. To ensemble them and train the ensemble methods we implemented (see above), we further nested a $3$-fold cross validation in each test fold, leaving $11\%$ of samples for ensemble training.

\par We assess the quality of our recommendations by comparing its results to the subset of ensemble methods implementable on TabArena, using a different diverse set of base learners to ensure our recommendations are model-agnostic: LightGBM, ExtraTrees, Modern NCA, RealMLP, Logistic Regression, ExplainableBM, xRFM and RealTabPFN 2.5. When multiple ensemble methods were recommended, we averaged their predictions. Stacking was not simulated.

\par To get close to real conditions, we performed a CASH optimization through the simulation of a $250$-iteration random search for every task and used the top $50$ model instances as base learners, reducing diversity but selecting only well performing base learners. Both this random search and the following experiments use ROC AUC, for the $22$ binary tasks, and negative log loss (NLL), for the $6$ multiclass tasks. We ensured no dataset from the exploratory study was present.

\par We had two testable questions: (1) is each ensemble method better than Single Best, and (2) are our recommendations better than the existing ensemble methods? We used Wilcoxon signed rank tests to test them, with a significance threshold $\alpha = 0.05$ and a Holm-Bonferroni correction~\cite{holm_simple_1979}.

\subsection{Results}

\begin{table}[ht]
    \centering
    \caption{Mean scores, mean ranks and task-wise rank MeAD of the ensemble methods over all tasks. Best results are underlined. The multiclass score (NLL) is a loss where smaller is better. }
    \label{tab:validation}
    \begin{tabular}{rcccccc}
    	\hline                 
    	&  \multicolumn{2}{c}{\textbf{All}}  &  \multicolumn{2}{c}{\textbf{Binary}}   &  \multicolumn{2}{c}{\textbf{Multiclass}} \\
    	& Rank & Win Rate   & \hspace{4mm}Rank\hspace{4mm}  &  Score  &   \hspace{4mm}Rank\hspace{4mm}  &  Score\\
    	\hline
        \textbf{Single Best}       & $4.4 \pm 0.7$ &              $0\%$ & $4.5 \pm 0.6$   &                 0.861 & $4.0 \pm 0.7$ &      0.290\\
        \hline
        \textbf{Recs}     & $\underline{2.3} \pm 0.9$ &              $29\%$ & $2.2 \pm 0.9$   &                 \underline{0.863} & $2.5 \pm 0.7$ &      0.279  \\
        \hline
        \textbf{Soft Voting}        & $2.5 \pm 1.0$ &              $25\%$ & $2.4 \pm 0.9$   &                 0.862 & $2.7 \pm 1.2$ &      \underline{0.277}\\
        \textbf{Blending}         & $2.5 \pm 1.3$ &              \underline{$32\%$} & $\underline{2.1} \pm 0.8$   &                 \underline{0.863} & $3.7 \pm 1.8$ &      0.298\\
        \textbf{Robust SV} & $3.6 \pm 1.1$ &              $14\%$ & $4.0 \pm 0.9$   &                 0.862 & $\underline{2.3} \pm 1.0$ &      \underline{0.277}\\
        \textbf{Hard Voting}           & $5.8 \pm 0.3$ &              $0\%$ & $5.8 \pm 0.3$   &                 0.809 & $5.8 \pm 0.3$ &      2.159 \\
        \hline
    \end{tabular}
\end{table}

\begin{table}[ht]
    \centering
    \caption{Corrected Wilcoxon signed rank test results of the ensemble methods. \ding{51} and \ding{55} means that the p-value is respectively below and above $0.05$.}
    \label{tab:tests}
    \begin{tabular}{rcc}
    	\hline                 
    	& \hspace{2mm}$>$ SB\hspace{2mm} & \hspace{2mm}$<$ Recs\hspace{2mm}\\
    	\hline
        \textbf{Single Best (SB)} & & \ding{51} \\
        \hline
        \textbf{Recs} & \ding{51} & \\
        \hline
        \textbf{Soft Voting} & \ding{51} & \ding{55}\\
        \textbf{Blending} & \ding{51} & \ding{55}\\
        \textbf{Robust SV} & \ding{51} & \ding{55}\\
        \textbf{Hard Voting} & \ding{55} & \ding{51} \\
        \hline
    \end{tabular}
\end{table}

\par \autoref{tab:validation} and \autoref{tab:tests} show the general results we obtained, as well as the results of the significance tests\footnote{Measures may be found at \doi{10.5281/zenodo.17986269}.}. Results differ from the previous experiment (Section~\ref{sec:exploratory}) in a few ways. Soft Voting and Robust SV, on one hand, show even stronger results on the multiclass tasks. Single Best, on the other hand, performed badly, being significantly worse than every method, except for Hard Voting.

\par Our set of recommendations (Recs) performed well, with slightly better results than Blending, especially in the multiclass case. Though it wasn't significantly better than the best methods, its results were more consistent than their's.

\par However, it may be necessary to update our recommendation $5.$ In our exploratory study, Single Best achieved a mean multiclass ROC AUC rank of $3.0$, comparable to Robust SV ($3.1$) and Stacking ($3.1$). However, this parity disappears in the validation study, where Single Best is significantly worse than all ensemble methods except Hard Voting. We therefore recommend Single Best only when a dominant base learner has been identified through cross-validation, as its otherwise weaker average behavior does not justify its use as a default.



\subsection{Stacking results}

\begin{table}[ht]
    \centering
    \caption{Results with the inclusion of Stacking.}
    \label{tab:validation_stacking}
    \begin{tabular}{rcccccc}
    	\hline                 
    	&  \multicolumn{2}{c}{\textbf{All}}  &  \multicolumn{2}{c}{\textbf{Binary}}   &  \multicolumn{2}{c}{\textbf{Multiclass}} \\
    	& Rank & Win Rate   & \hspace{4mm}Rank\hspace{4mm}  &  Score  &   \hspace{4mm}Rank\hspace{4mm}  &  Score\\
        \hline
         \textbf{Single Best}       & $5.0 \pm 0.6$ &              $0\%$ & $5.2 \pm 0.6$   &               0.861 & $4.2 \pm 0.6$ &      0.290 \\
        \hline
         \textbf{Recs}      & $\underline{2.7} \pm 0.7$ &              $3\%$ & $2.7 \pm 0.7$   &               \underline{0.863} & $2.7 \pm 0.8$ &      0.279 \\
         \hline
         \textbf{Soft Voting}        & $2.9 \pm 1.4$ &              $25\%$ & $3.0 \pm 1.3$   &               0.862 & $2.8 \pm 1.4$ &      \underline{0.277} \\
          \textbf{Blending}         & $2.8 \pm 1.3$ &              \underline{$29\%$} & $\underline{2.5} \pm 1.1$   &               \underline{0.863} & $3.8 \pm 1.6$ &      0.298 \\
         \textbf{Robust SV} & $4.1 \pm 1.6$ &              $14\%$ & $4.6 \pm 1.3$   &               0.862 & $\underline{2.5} \pm 1.2$ &      \underline{0.277} \\
         \textbf{Hard Voting}           & $6.7 \pm 0.5$ &              $0\%$ & $6.8 \pm 0.4$   &               0.809 & $6.5 \pm 0.7$ &      2.159 \\
         \hline
          \textbf{Stacking}         & $3.7 \pm 2.0$ &              \underline{$29\%$} & $3.2 \pm 1.8$   &               \underline{0.863} & $5.5 \pm 1.5$ &      2.276 \\
        \hline
        \end{tabular}
\end{table}

\par Implementing Stacking in TabArena required more adaptations than the other methods. We had to use two independent $3$-fold cross-validations for the base learners and for the meta-dataset. However, data leaked. Specifically, the learners used to create the meta-dataset (but not for predictions) were each trained on part of the test set. For this reason, we separated its results from the rest.

\par \autoref{tab:validation_stacking} shows the results after including Stacking. Despite data leaks benefiting it, the performance Stacking showed here were considerably more mixed than in the previous study. While it was the best method in $8$ out of $28$ cases, it was also the worst or second worst method in another $10$ out of $28$ cases.

\par Surprisingly, all of Stacking's wins on binary tasks occurred on datasets where our recommendations would have otherwise won. Stacking seems at its best when the rest of our recommendations are already strong, and generally worse on other tasks. Considering its mediocre results here, it might be wise to avoid using it when combining recommendations by averaging them.








\section{Conclusion}
\label{sec:conclusion}

\par In this work, we assessed the performance of several heterogeneous ensemble models and introduced new recommendations for their use, which we validated empirically. To enable this validation, we extended TabArena to allow the use of different ensemble methods.

\par Specifically, we noted that Hard Voting shouldn't be used for probabilistic classification. Stacking is a strong ensemble method that suffers from inconsistencies, particularly in the multiclass case and when base learners have correlated results. Blending is a version of Stacking with lower performance in the best case but with less inconsistencies. Interestingly, they perform badly in different cases. Soft Voting is a good alternative on multiclass cases, although it suffers from bad base learners; Robust SV suffers slightly less from this problem. Single Best is a good alternative if a base learner is much stronger than the others; otherwise, it is significantly worse than ensemble methods.

\par Further work on optimizations for ensemble models such as pruning and weight optimization techniques would compliment the current work. A larger-scale comparison using the base learners of the second experiment  with all methods of the first experiment would also allow a more thorough comparison.

\bibliographystyle{splncs04}
\bibliography{bibliography}
\clearpage
\appendix

\section{Ensemble methods}

\par We briefly describe the ensemble methods used and the implementation for the exploratory experiments.

\subsection{Description}

\par We use the following notations for all definitions. Let $\mathcal{X}$ be the feature space for a classification problem with a set of classes $C = (c_k)_{k \leq n_C}$. Let $B = (b_i)_{i \leq n_B}$ be a set of trained base learners whose respective error is $\epsilon^{(i)}_{A} \in 
\mathds{R}^{+}, \forall A \subset \mathcal{X}$, whose class predictor is $h^{(i)}: \mathcal{X} \mapsto C$, and whose probability predictor is $f^{(i)}: \mathcal{X} \mapsto \Delta^{n_C}$, with $\Delta^{n_C}$ the standard $n_C$-simplex. $\forall i \leq n_B, \forall k \leq n_C,$ let $f_k^{(i)}: \mathcal{X} \mapsto [0;1]$ be the probability predictor for $c_k$, such that  $f^{(i)} = [f_k^{(i)}]_{k=1}^{n_C}$. $\forall i, j \in \mathds{N}^2$, let $\delta_{ij}$ be the Kronecker delta. Let $x \in \mathcal{X}$ be an unknown sample.

\medskip

\par \textbf{Soft Voting} uses the arithmetic mean of the base learners' class probabilities for its predictions. Fumera and Roli~\cite{fumera_soft_voting_2005} analyzed its decision boundary~\cite{tumer_ensemble_1996}, and showed the method is able to reduce its base learners' added error - the non-intrinsic part of the error - by a factor equal to the size of its pool in the best case. Its probability predictor $f^{sv}$ is defined as 

\begin{equation*}
    f^{sv}(x) \triangleq [\operatornamewithlimits{mean}_{i \leq n_B}({f_k^{(i)}(x)})]_{k=1}^{n_C} = \frac{1}{n_B}\sum_{i = 1}^{n_B}{f^{(i)}(x)}~.
\end{equation*}
    
\par \textbf{Robust Soft Voting} uses the median of the base learners' class probabilities for its predictions, normalized to remain a probability. It's a variant of Soft Voting that uses a more robust measure of central tendency. Its probability predictor $f^{rsv}$ is defined $\forall x \in \mathcal{X}$ as the only value in $\Delta^{n_C}$ that  satisfies

\begin{equation*}
    f^{rsv}(x) \propto [\operatornamewithlimits{median}_{i \leq n_B}({f_k^{(i)}(x)})]_{k=1}^{n_C}~.
\end{equation*}

\par \textbf{Hard Voting} or Plurality Voting uses its base learners' predictions as votes and chooses the class with the most votes. The jury theorem~\cite{condorcet_essai_1785} states, under several strict hypotheses, that its accuracy should increase with the size of the pool and converge to the best possible predictor with an infinite pool for binary classification. Its class predictor $h^{rsv}$ and probability predictor $f^{rsv}$ are defined as

\begin{equation*}
    h^{hv}(x) = \argmax_{c \in C}(\left|\{i \leq n_B, h^{(i)}(x) = c\}\right|)~\text{and}
\end{equation*}
\begin{equation*}
    f^{hv}(x) = [\frac{1}{n_B}\left|\{i \leq n_B, h^{(i)}(x) = c_k\}\right|]_{k=1}^{n_C}~.
\end{equation*}

\par \textbf{Stacking} and \textbf{Blending} use the predictions of its base learners as features to train a logistic regression. It leads to the creation of an unconstrained weighted soft-voting model by class with weights $(\beta^{(i)})_{1 \leq i \leq n_B} \in (\mathds{R}^{n_C})^{n_B}$ as well as an intercept $\beta^{(0)} \in \mathds{R}^{n_C}$, both given to the non-linear bijective logistic function $\sigma$. If properly optimized and in the best case, a weighted soft-voting model should theoretically completely remove its base learners' added error~\cite{fumera_soft_voting_2005}. Its probability predictor $f^{st}$ is defined as 

\begin{equation*}
    f^{st}(x) = [\sigma(\beta^{(0)}_k + \sum_{i = 1}^{n_B}{\beta^{(i)}_kf^{(i)}(x)})]_{k=1}^{n_C}~.
\end{equation*}

\par \textbf{Dynamic Classifier Selection} first trains the base learners on all samples, then partitions the feature space into a disjoint union $\mathcal{X} = \bigsqcup_{j = 1}^{n_X}{\mathcal{X}_j}$ and selects the best performing base learner on each part. During prediction, only the base learner corresponding to the part the sample falls in is used for prediction. As the best performing base learner is selected for each part, the resulting ensemble model is expected to perform better than any of its base learners. Its probability predictor $f^{dcs}$ is defined $\forall j \leq n_X$, with $a_j \triangleq \argmin_{l \leq n_B}{\epsilon^{(l)}_{\mathcal{X}_j}}$, if $x \in \mathcal{X}_j$, by

\begin{equation*}
    f^{dcs}(x) \triangleq {\sum_{i = 1}^{n_B}{\mathds\delta_{a_j i}f^{(i)}(x)}}~.
\end{equation*}

\par \textbf{Mixture of Experts} first partitions the feature space, then trains different instances of all base learner on each part $\forall j \leq n_X, f^{(i, j)} : \mathcal{X}_j \mapsto \Delta^{n_C}$ , and selects the best performing base learner on each part. Similarly to spline interpolation, base learners are expected to model simple parts of the true distribution. Its probability predictor $f^{moe}$ is defined $\forall j \leq n_X$, with $a_j \triangleq \argmin_l{\epsilon^{(l)}_{\mathcal{X}_j}}$, if $x \in \mathcal{X}_j$, by

\begin{equation*}
    f^{moe}(x) \triangleq {\sum_{i = 1}^{n_B}{\mathds\delta_{a_j i}f^{(i, j)}(x)}}~.
\end{equation*}

\subsection{Exploratory experiments}

\par While we created custom ensemble methods for the validation experiments, we were able to use and customize existing implementations for the exploratory experiments.

\begin{description}
    \item \textbf{Soft Voting} (\texttt{mlxtend}): \texttt{EnsembleVoteClassifier}
    \begin{itemize}
        \item \texttt{voting = 'soft'}
    \end{itemize}
    \vspace{8pt}
    \item \textbf{Hard Voting} (\texttt{mlxtend / custom}): Modified \texttt{EnsembleVoteClassifier} with a custom \texttt{predict\_proba}
    \begin{itemize}
        \item \texttt{voting = 'hard'}
    \end{itemize}
    \vspace{8pt}
    \item \textbf{Bagging} (\texttt{scikit-learn / custom}): Modified \texttt{BaggingClassifier} with the possibility of using heterogeneous base learners
    \vspace{8pt}
    \item \textbf{Stacking} (\texttt{scikit-learn}): \texttt{StackingClassifier}
    \begin{itemize}
        \item \texttt{final\_estimator = LogisticRegression}
    \end{itemize}
    \vspace{8pt}
    \item \textbf{Blending} (\texttt{scikit-learn}): \texttt{StackingClassifier}
    \begin{itemize}
        \item \texttt{final\_estimator = LogisticRegression}
        \item \texttt{cv = 'prefit'}
    \end{itemize}
    \vspace{8pt}
    \item \textbf{Cluster DCS} (\texttt{custom / umap}): Custom cluster DCS method using UMAP~\cite{mcinnes_umap_2020} for $3$ dimensions and HDBSCAN~\cite{mcinnes2017hdbscan} with their recommended hyperparameters~\cite{mcinnes_using_2018}.
    \vspace{8pt}
    \item \textbf{Cluster MoE} (\texttt{custom / scikit-learn}): Custom cluster MoE method using $5$-means clustering.
    \vspace{8pt}
\end{description}

\section{Base learners}

\subsection{Exploratory experiments}

\par We note the hyperparameters used for each base learner in the exploratory experiment. Hyperparameters that aren't explicitly given were left to their default values in their given library.

\par The libraries we used were \texttt{scikit-learn 1.7.1}~\cite{pedregosa_scikit-learn_2011} and \texttt{xgboost 3.0.4}~\cite{chen_xgboost_2016}.

\begin{description}
    \item \textbf{Most Common} (\texttt{custom}): Custom model that always predicts the most common class seen in training, and gives every class the probability corresponding to its frequency in the training set.
    \vspace{8pt}
    \item \textbf{Linear SVM} (\texttt{scikit-learn}): \texttt{SVC}
    \begin{itemize}
        \item \texttt{kernel = 'linear'}
        \item \texttt{probability = True}
        \item \texttt{max\_iter = 1 000 000}
    \end{itemize}
    \vspace{8pt}
    \item \textbf{Shallow Decision Tree} (\texttt{scikit-learn}): \texttt{DecisionTreeClassifier}
    \begin{itemize}
        \item \texttt{max\_depth = 3}
    \end{itemize}
    \vspace{8pt}
    \item \textbf{Gaussian Naïve Bayes} (\texttt{scikit-learn}): \texttt{GaussianNB}
    \vspace{8pt}
    \item \textbf{Logistic Regression} (\texttt{scikit-learn}): \texttt{LogisticRegression}
    \begin{itemize}
        \item \texttt{penalty = 'l2'}
        \item \texttt{solver = 'lbgfs'}
    \end{itemize}
\end{description}

\begin{description}
    
    \item \textbf{Decision Tree} (\texttt{scikit-learn}): \texttt{DecisionTreeClassifier}
    \vspace{8pt}
    \item \textbf{SVM} (\texttt{scikit-learn}): \texttt{SVC}
    \begin{itemize}
        \item \texttt{kernel = 'rbf'}
        \item \texttt{probability = True}
        \item \texttt{max\_iter = 1 000 000}
    \end{itemize}
    \vspace{8pt}
    \item \textbf{Random Forest} (\texttt{scikit-learn}): \texttt{RandomForestClassifier}
    \begin{itemize}
        \item \texttt{n\_estimators = 100}
    \end{itemize}
    \vspace{8pt}
    \item \textbf{Multilayer Perceptron $l \times n$} (\texttt{scikit-learn}): \texttt{MLPClassifier}:
    \begin{itemize}
        \item \texttt{hidden\_layer\_sizes = [n] * l}
        \item \texttt{tol = 1e-4}
    \end{itemize}
    \vspace{8pt}
    \item \textbf{XGBoost} (\texttt{xgboost}): \texttt{XGBClassifier}
\end{description}

\par Note that the logistic regression used as a final estimator in Blending and Stacking uses the same hyperparameters described above. 

\subsection{Validation experiment}

\par For the validation experiments, we used TabArena's results~\cite{erickson_tabarena_2025} and their hyperparameter ranges. For most base learners, they are noted in the appendix C of the TabArena paper. The ranges for xRFM and RealTabPFN are not noted; we show their hyperparameter values and ranges, taken from TabArena's code, below: 


\begin{description}
    \item \textbf{xRFM} (\texttt{tabarena}): \texttt{XRFMModel}
    \begin{itemize}
        \item \texttt{standardize\_cats = False}
        \item \texttt{bandwidth\_mode = 'constant'}
        \item \texttt{early\_stop\_multiplier = 1.1}
        \item \texttt{solver = 'solve'}
        \item \texttt{classification\_mode = 'prevalence'}

        \item \texttt{bandwidth: LogUniform[.5, 200]}
        \item \texttt{diag: Choice[False, True]}
        \item \texttt{exponent: Choice[.7, 1.4]}
        \item \texttt{p\_interp: Choice[0, .8]}
        \item \texttt{kernel: Choice['lpq, kermac', 'l2']}
        \item \texttt{reg: LogUniform[1e-6, 1.]}
    \end{itemize}
    \vspace{8pt}
    \item \textbf{RealTabPFN2.5} (\texttt{tabarena}): \texttt{RealTabPFNv25Model}
    \begin{itemize}
        \item \texttt{softmax\_temperature: Choice[.25,.5,.6,.7,.8,.9,1.,1.25,1.5]}
        \item \texttt{balance\_probabilities: Choice[True, False]}
        \item \texttt{inference\_config/OUTLIER\_REMOVAL\_STD: Choice[3,6,12]}
        \item \texttt{inference\_config/POLYNOMIAL\_FEATURES: Choice['no', 25]}
        \item \texttt{inference\_config/REGRESSION\_Y\_PREPROCESS\_TRANSFORMS: Choice[}
        \begin{description}
            \item \texttt{[None],}
            \item \texttt{[None, 'safepower'],}
            \item \texttt{['safepower'],}
            \item \texttt{['kdi\_alpha\_0.3'],}
            \item \texttt{['kdi\_alpha\_1.0'],}
            \item \texttt{['kdi\_alpha\_3.0'],}
            \item \texttt{['quantile\_uni'],}
        \end{description}
        \texttt{]}
        \item \texttt{preprocessing/scaling: Choice[}
        \begin{description}
            \item \texttt{['none'],}
            \item \texttt{['quantile\_uni\_coarse'],}
            \item \texttt{['quantile\_norm\_coarse'],}
            \item \texttt{['kdi\_uni'],}
            \item \texttt{['kdi\_alpha\_0.3'],}
            \item \texttt{['kdi\_alpha\_3.0'],}
            \item \texttt{['safepower', 'quantile\_uni'],}
            \item \texttt{['none', 'quantile\_uni\_coarse'],}
            \item \texttt{['squashing\_scaler\_default', 'quantile\_uni\_coarse'],}
            \item \texttt{['squashing\_scaler\_default'],}
        \end{description}
        \texttt{]},
        \item \texttt{preprocessing/categoricals: Choice['numeric','onehot','none']}
        \item \texttt{preprocessing/append\_original: Choice[True, False]}
        \item \texttt{preprocessing/global: Choice[one, 'svd', 'svd\_quarter\_components']}
    \end{itemize}
\end{description}

\end{document}